\documentclass[sigconf,nonacm]{acmart} 

\acmSubmissionID{415}

\usepackage[inline]{enumitem}
\usepackage{stfloats} 
\usepackage{amsmath}
\usepackage{multirow}
\usepackage{subcaption}
\usepackage{xspace}
\usepackage{microtype}

\copyrightyear{2024} 
\acmYear{2024} 
\setcopyright{acmlicensed}\acmConference[SIGGRAPH Conference Papers '24]{Special Interest Group on Computer Graphics and Interactive Techniques Conference Conference Papers '24}{July 27-August 1, 2024}{Denver, CO, USA}
\acmBooktitle{Special Interest Group on Computer Graphics and Interactive Techniques Conference Conference Papers '24 (SIGGRAPH Conference Papers '24), July 27-August 1, 2024, Denver, CO, USA}
\acmDOI{10.1145/3641519.3657422}
\acmISBN{979-8-4007-0525-0/24/07}

\newcommand{\sysname}{LGTM\xspace}
\definecolor{blue}{rgb}{0,0,1}
\definecolor{red}{rgb}{1,0,0}
\definecolor{green}{rgb}{0,.5,0}
\definecolor{orange}{rgb}{0.75, 0.4, 0}
\definecolor{teal}{rgb}{0.0, 0.4, 0.4}
\definecolor{purple}{rgb}{0.65,0,0.65}

\newcommand{\revision}[1]{#1}

\newcommand{\ie}{{\it i.e.}}

\begin{document}

\title{LGTM: Local-to-Global Text-Driven Human Motion \\ Diffusion Model}

\author{Haowen Sun}
\orcid{0000-0002-6450-7746}
\affiliation{%
    \institution{Shenzhen University}
    \country{China}
}
\email{zover.v@gmail.com}

\author{Ruikun Zheng}
\affiliation{%
    \institution{Shenzhen University}
    \country{China}
}
\email{zhengzrk@gmail.com}

\author{Haibin Huang}
\orcid{0000-0002-7787-6428}
\affiliation{
    \institution{Kuaishou Technology}
    \country{China}
}
\email{jackiehuanghaibin@gmail.com}

\author{Chongyang Ma}
\orcid{0000-0002-8243-9513}
\affiliation{
    \institution{ByteDance Inc.}
    \country{USA}
}
\email{chongyangm@gmail.com}

\author{Hui Huang}
\orcid{0000-0003-3212-0544}
\affiliation{%
    \institution{Shenzhen University}
    \country{China}
}
\email{hhzhiyan@gmail.com}

\author{Ruizhen Hu}
\authornote{Corresponding author: Ruizhen Hu (\href{mailto:ruizhen.hu@gmail.com}{ruizhen.hu@gmail.com})}
\orcid{0000-0002-6798-0336}
\affiliation{%
    \institution{Shenzhen University}
    \country{China}
}
\email{ruizhen.hu@gmail.com}

\renewcommand{\shorttitle}{LGTM: Local-to-Global Text-Driven Human Motion Diffusion Model}


\begin{abstract}
    In this paper, we introduce LGTM,  a novel \textbf{L}ocal-to-\textbf{G}lobal pipeline for \textbf{T}ext-to-\textbf{M}otion generation.  LGTM utilizes a diffusion-based architecture and aims to address the challenge of accurately translating textual descriptions into semantically coherent human motion in computer animation. Specifically, traditional methods often struggle with semantic discrepancies, particularly in aligning specific motions to the correct body parts. To address this issue, we propose a two-stage pipeline to overcome this challenge: it first employs large language models (LLMs) to decompose global motion descriptions into part-specific narratives, which are then processed by independent body-part motion encoders to ensure precise local semantic alignment. Finally, an attention-based full-body optimizer refines the motion generation results and guarantees the overall coherence. Our experiments demonstrate that LGTM gains significant improvements in generating locally accurate, semantically-aligned human motion, marking a notable advancement in text-to-motion applications. \revision{Code and data for this paper are available at \url{https://github.com/L-Sun/LGTM}}
\end{abstract}



\begin{teaserfigure}
    \includegraphics[width=\textwidth]{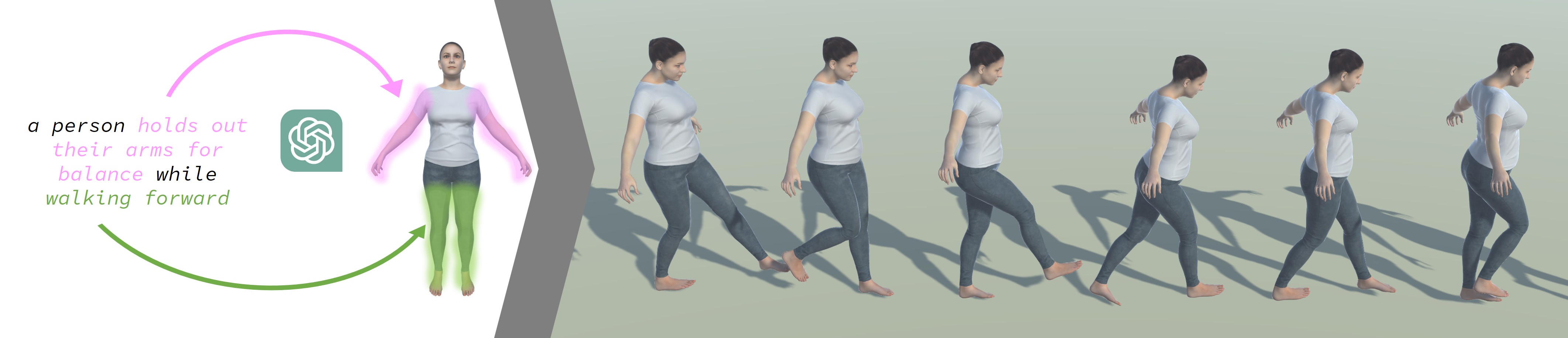}
    \caption{LGTM decomposes input text description at the part level using LLMs and generates output motion from local to global with both part description and full-body descriptions.}
    \label{fig:teaser}
\end{teaserfigure}


\maketitle

\section{Introduction}
In this paper, we address the problem of text-to-motion, \ie, given a textual description of movements for a character, we aim to automatically generate plausible and realistic 3D human motions. The successful automation of this process holds significant potential for a variety of downstream applications, including the creation of content for augmented and virtual reality environments, advancements in robotics, and enhancements in human-machine interactions \cite{scanlon2023ChatGPTDigitalForensic,lan2023ApplicationChatGPTBasedDigital,zhaoMetaversePerspectivesGraphics2022,chenEfficientPhotorealistic3D2021}.

As a longstanding challenge at the confluence of natural language processing, machine learning, and computer graphics, text-to-motion generation has garnered significant attention in recent research~\cite{petrovich2022TEMOSGeneratingDiverse,tevet2022MotionCLIPExposingHuman, jiang2023MotionGPTHumanMotion}. The advent of diffusion models, as highlighted in various studies~\cite{rombach2022HighResolutionImageSynthesisa, poole2022DreamFusionTextto3DUsing, alexanderson2023ListenDenoiseAction}, has propelled notable advancements in this field~\cite{tevet2022HumanMotionDiffusion}. Despite these strides, the task of generating motions that are both locally semantic accurate and globally coherent from textual descriptions remains a formidable hurdle. Current methods often face difficulties in effectively capturing the nuanced local semantics embedded in motion descriptions and in producing motions that align accurately with these semantic cues.

In particular, existing approaches in text-to-motion synthesis often encounter issues such as local semantic leakage and missing elements~\cite{tevet2022HumanMotionDiffusion, chen2023ExecutingYourCommands}. For instance, when prompted with a description like ``a man kicks something with his left leg'', these methods might erroneously generate a motion that corresponds to a ``right kick''. Similarly, prompts involving complex actions requiring coordination of multiple body parts frequently result in motions with certain parts omitted.
Our observations reveal two primary shortcomings in these methods.
Firstly, most existing techniques utilize a single global text descriptor for all local body motions. This approach requires the network to learn the association between local motion semantics and respective body parts from a unified global text source. This process proves challenging, especially when the textual content bears similarity across different body parts, leading to difficulties in differentiating specific actions for each part.
Secondly, the text encoders used in these methods exhibit limited effectiveness in encoding motion-related text. This limitation is apparent in the high feature similarity observed among different motion texts, as detailed in recent studies \cite{petrovich2023TMRTexttoMotionRetrieval}. This homogeneity in encoded text features further exacerbates the network's struggle to discern and accurately represent subtle variations in local textual semantics.

Towards this end, we present a novel diffusion-based text-to-motion generation architecture, \sysname, adept at producing motions that are both in alignment with textual descriptions and precise in local semantic accuracy. \sysname operates through a local-to-global approach, structured in two main stages. The first stage implements an efficient strategy to tackle the issue of local semantic accuracy. Here, we introduce a partition module that employs large language models (LLMs) to dissect global motion descriptions into narratives specific to each body part. Subsequently, dedicated body-part motion encoders independently process these part-specific narratives. This focused approach effectively circumvents local semantic inaccuracies by reducing redundant information and preventing semantic leakage, thus maintaining a sharp focus on relevant local semantics. However, as each body-part motion encoder functions independently, without awareness of other parts' movements, it is imperative to synchronize these individual motions to avoid full-body coordination issues. To address this, the second stage of \sysname introduces an attention-based full-body optimizer. This component is specifically designed to facilitate the integration of information among different body parts, ensuring that the overall motion is not only locally precise but also globally coherent and fluid.

To evaluate the effectiveness of \sysname, we further conduct experiments on text-driven motion generation and provide both quantitative and qualitative results. Our experiments show that our proposed \sysname can generate faithful motions that better align with the input text both locally and globally, and outperform state-of-the-art methods.

To summarize, our contributions are as follows:
\begin{itemize}[leftmargin=*]
    \item We present \sysname, a novel diffusion-based architecture that translate textual descriptions into accurate and coherent human motions, marking a significant improvement over previous text-to-motion approaches.

    \item  \sysname introduces a unique partition module that utilizes LLMs to decompose complex motion descriptions into part-specific narratives. This significantly enhances local semantic accuracy in motion generation.

    \item Our experiments demonstrate the effective integration of independent body-part motion encoders with an attention-based full-body optimizer, ensuring both local precision and global coherence in generated motions, providing a promising improvement for text-to-motion generation.
\end{itemize}

\section{Related Work}
The generation of motion sequences is a longstanding challenge within the domain of computer graphics, where the objective is to produce a series of motion frames guided by conditional control signals. Given that our approach is centered on body-partition-based text-to-motion synthesis, we explore relevant literature across two primary aspects: body partition modeling and text-to-motion generation.

\paragraph{Part-based motion modeling.}
Partitioning the human body into distinct segments facilitates the control of motion synthesis at a more granular level, allowing for localized adjustments.

Several studies have explored the concept of combining motions of individual body parts to synthesize novel motions. \cite{hecker2008RealtimeMotionRetargeting} introduced a retargeting algorithm that composes motions at the level of individual body parts to generate diverse character animations. \cite{jang2008EnrichingMotionDatabase}  divided motions into upper and lower body segments, merging them through an algorithm to augment their motion database. \cite{soga2016BodypartMotionSynthesis} synthesized dance motions from existing datasets by focusing on body partitions. \cite{jang2022MotionPuzzleArbitrary} performed style transfer at the part level, utilizing a graph convolutional network to assemble different body part motions into new, coherent sequences, preserving local styles while transferring them to specific body parts without compromising the integrity of other parts or the entire body. However, these methods rely on pre-existing motion data, and hence are more accurately described as synthesis rather than generation.

For more detailed local control, \cite{starke2020LocalMotionPhases} proposed a local phase model based on body partitions used to generate basketball player movements, achieving higher local fidelity compared to global phase approaches \cite{zhang2018ModeadaptiveNeuralNetworks, starke2019NeuralStateMachine}. \cite{starke2021NeuralAnimationLayering}  introduced a neural animation layering technique that combines trajectories of different body parts produced by control modules, providing animators with more granular control and enabling the creation of high-quality motion. \cite{lee2022LearningVirtualChimeras} developed an algorithm for reassembling physically-based part motions, allowing the combination of partial movements from characters with varying skeletal structures. By operating in a physically simulated virtual environment, they employed part-wise timewarping and optimization-based assembly to ensure improved spatial and temporal alignment.  \cite{bae2023PMPLearningPhysically} utilized part-wise motion discriminators to enhance motion variety and a global control policy to maintain the physical realism of the movements.

\paragraph{Text-to-motion generation.}
Text provides a user-friendly interface for directing motion generation due to its ease of use and editing capabilities. However, a significant challenge arises from the difficulty in precisely controlling the outcome of the generated motion through text. In this subsection, we examine text-to-motion generation techniques and identify their limitations.

Certain text-to-motion approaches are founded on the encoder-decoder architecture and focus on aligning modalities within a unified latent space. \cite{ahuja2019Language2poseNaturalLanguage} trained their network by alternating between encoding motions and texts, then decoding them back into motion, thereby implicitly aligning the two modalities. \cite{ghosh2021SynthesisCompositionalAnimations,petrovich2022TEMOSGeneratingDiverse} encoded text and motion concurrently and decoded them into motion, employing additional loss functions to bring the modalities closer within the latent space.  These methods struggle with generating motions from lengthy textual descriptions. \cite{athanasiou2022TEACHTemporalAction} tackled long motion generation by producing short motion clips in an auto-regressive fashion, but this requires manual segmentation of long textual descriptions into shorter segments and specification of action duration. To utilize visual priors, \cite{tevet2022MotionCLIPExposingHuman} employed a frozen CLIP~\cite{radford2021LearningTransferableVisual} text encoder to encode motion descriptions and aligned the motion latent space with that of CLIP. Nevertheless, the images used for alignment, rendered from random motion frames, can confuse the network when the frames are not representative. Moreover, \cite{petrovich2023TMRTexttoMotionRetrieval} observed that motion descriptions tend to cluster closely in the CLIP latent space, as the distribution of motion-related text is narrower than that of the broader text datasets used to train CLIP.

\begin{figure*}
    \centering
    \includegraphics[width=\textwidth]{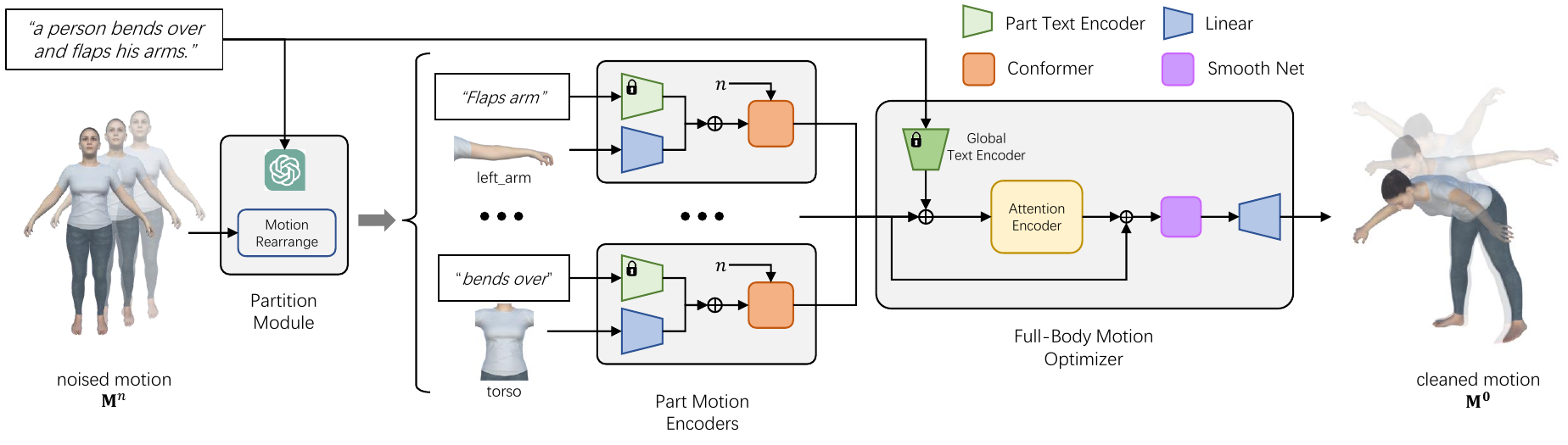}
    \caption[]{Overview of our \sysname  framework, which consists of three major components.
        \begin{enumerate*}
            \item The partition module utilizes ChatGPT to deconstruct motion descriptions $T$ into body part level text $T_\mathrm{part}$, and decomposes full-body motion $\mathbf{M}$ to body part motion $\mathbf{M}_\mathrm{part}$;
            \item The part motion encoders encodes part-level motions with corresponding part-level text independently and a diffusion time step $n$;
            \item The full-body motion optimizer utilizes an attention module to optimize fused body part motion with full-body text semantic.
        \end{enumerate*}}
    \label{fig:overview}
\end{figure*}

Recent developments in neural diffusion models for image generation have inspired text-to-motion methods that leverage these models to achieve superior quality.  \cite{tevet2022HumanMotionDiffusion, zhang2022MotionDiffuseTextDrivenHuman} utilized Transformer to denoise motion conditioned on text. \cite{chen2023ExecutingYourCommandsa} introduced a U-Net-based DDIM generative model to denoise motion in latent space, resulting in expedited generation. However, these methods lack the ability to control localized motion generation through masking. Additionally, they struggle to learning correct mapping of the local semantics because all body parts share the same textual information, which potentially lead to semantically mismatched part motions.

An alternative approach to motion generation involves processing motion in a discrete space through token prediction \cite{guo2022TM2TStochasticTokenized, jiang2023MotionGPTHumanMotion, yaoMoConVQUnifiedPhysicsBased}.  But the limitations of these works are that the expressive capacity of the codebook can restrict the diversity of the generated motions, potentially causing the text input to be mapped to unintended motions.

The challenges in controlling local motion semantics stem from:
\begin{enumerate*}
    \item the sharing of textual information across all body parts, and
    \item the difficulty networks face in distinguishing text latent codes encoded by CLIP.
\end{enumerate*}
These factors contribute to the difficulty of achieving precise local semantic control in motion generation, leading to issues such as semantic leakage.

Drawing inspiration from the technological advancements and challenges identified in prior research, we propose a novel framework that combines body-part partitioning with independent local motion semantic injection and a global semantic joint optimization strategy. This framework is designed to enhance the fidelity and controllability of text-to-motion synthesis, addressing the need for more nuanced and accurate motion generation.

\section{Method}

In this section, we delve into the specifics of \sysname, as illustrated in Figure~\ref{fig:overview}. \sysname is structured as a local-to-global generation framework that initially creates local, part-level motion, followed by a global fusion and optimization process to produce the final full-body motion. At its core, \sysname operates by subdividing the full-body text and motion spaces into body-part-specific subspaces. Such subdivision is adeptly handled by a dedicated Partition Module.

For each of these subspaces, we have developed specialized part motion encoders. These encoders are trained to learn independently a series of mappings between part-level motions and part-level text. This strategy effectively mitigates the issues of incorrect local semantic mapping seen in previous methods. Following the localized encoding, \sysname introduces a full-body motion optimizer to establish correlations among the various subspaces and ensure the consistency and coherence of the final full-body motion.
Below, we provide a detailed explanation of the functionalities and details of each module in \sysname.

\subsection{Preliminary: Human Motion Diffusion Model}

\paragraph{Input representation.}
\revision{We define the input pair for our method as $\left(\mathbf{M}, T\right)$, where $\mathbf{M}$ represents full-body motion data and $T$ denotes the raw full-body text description.}
Specifically, we use the HumanML3D representation proposed by \cite{guo2022GeneratingDiverseNatural} as our motion data representation, which is calculated from the SMPL motion data~\cite{loper2015SMPLSkinnedMultiperson} and includes redundant motion features that are helpful for network training. A full-body motion data $\mathbf{M}$ contains $F$ frames and $J=22$ joints. Specifically, we denote $\mathbf{M} = [\dot{\mathbf{r}}_\mathrm{root}, v_\mathrm{root}, h, \mathbf{p}, \mathbf{r},  \mathbf{v}, \mathbf{c}]$, where $ \dot{\mathbf{r}}_\mathrm{root} \in \mathbb{R}^{F \times 1}$, $ v_\mathrm{root} \in \mathbb{R}^{F \times 2}$ and $ h \in \mathbb{R}^{F \times 1}$ are the angular velocity around y-axis, linear velocity on x-z plane, and height of the root joint,
$ \mathbf{p} \in \mathbb{R}^{F \times (J-1) \times 3}$ and $ \mathbf{r} \in \mathbb{R}^{F \times (J-1) \times 6} $ are local position and 6D rotation \cite{zhang2018ModeadaptiveNeuralNetworks}  of all joints except root joint, $ \mathbf{v} \in \mathbb{R}^{F \times J \times 3} $ is the local velocity of all joints, and  $\mathbf{c} \in \mathbb{R}^{F \times 4}$ is the contact signal of feet.

\paragraph{Diffusion model.}
\revision{Our method is built upon a text-conditional diffusion model. In the training stage, this model adds noise to a clean motion $\mathbf{{M}}$ following the Markov process and trains a network to predict the added noise with an L2 loss. In the sampling stage, this model gradually reduces noise from a purely noised motion $\mathbf{M}^n$ with the predicted noise. We use the DDIM \cite{song2022DenoisingDiffusionImplicit} as our diffusion model to accelerate the sampling process. More details is provided in the supplementary material.
}

\subsection{Partition Module}
\revision{The Partition Module is designed to inject local semantics into each body part for Part Motion Encoders. In practice, an input pair $\left(\mathbf{M}, T\right)$ is divided into six parts, including \textit{head}, \textit{left arm}, \textit{right arm}, \textit{torso}, \textit{left leg}, and \textit{right leg}.

    The motion $\mathbf{M}$ is decomposed as follows:
}
\begin{equation*}
    \begin{aligned}
        \mathbf{M}_\mathrm{head}       & = \left[ \mathbf{p}_\mathrm{head}, \mathbf{r}_\mathrm{head}, \mathbf{v}_\mathrm{head}  \right]                                                        \in \mathbb{R}^{F \times 24}  \\
        \mathbf{M}_\mathrm{left\_arm}  & = \left[ \mathbf{p}_\mathrm{left\_arm}, \mathbf{r}_\mathrm{left\_arm}, \mathbf{v}_\mathrm{left\_arm}  \right]                                         \in \mathbb{R}^{F \times 48}  \\
        \mathbf{M}_\mathrm{right\_arm} & = \left[ \mathbf{p}_\mathrm{right\_arm}, \mathbf{r}_\mathrm{right\_arm}, \mathbf{v}_\mathrm{right\_arm}  \right]                                      \in \mathbb{R}^{F \times 48}  \\
        \mathbf{M}_\mathrm{torso}      & = \left[ \mathbf{p}_\mathrm{torso}, \mathbf{r}_\mathrm{torso}, \mathbf{v}_\mathrm{torso}, \dot{\mathbf{r}}_\mathrm{root}, v_\mathrm{root}, h  \right] \in \mathbb{R}^{F \times 43}  \\
        \mathbf{M}_\mathrm{left\_leg}  & = \left[ \mathbf{p}_\mathrm{left\_leg}, \mathbf{r}_\mathrm{left\_leg}, \mathbf{v}_\mathrm{left\_leg}, \mathbf{c}_\mathrm{left\_leg}  \right]          \in \mathbb{R}^{F \times 50}  \\
        \mathbf{M}_\mathrm{right\_leg} & = \left[ \mathbf{p}_\mathrm{right\_leg}, \mathbf{r}_\mathrm{right\_leg}, \mathbf{v}_\mathrm{right\_leg},  \mathbf{c}_\mathrm{right\_leg} \right]      \in \mathbb{R}^{F \times 50}, \\
    \end{aligned}
\end{equation*}
where the subscript indicates where the feature from. For example, $\mathbf{p}_\mathrm{right\_leg}$ includes all local positions of joints from the \textit{right leg}.

For the motion description $T$, we leverage the knowledge inference capabilities of LLMs to decompose it into six parts: $T_\mathrm{head}$, $T_\mathrm{left\_arm}$, $T_\mathrm{right\_arm}$, $T_\mathrm{torso}$, $T_\mathrm{left\_leg}$ and $T_\mathrm{right\_leg}$ using crafted prompts. The prompt includes three sections: task definition, output requirements, and some output examples. The task definition instructs LLMs to extract principal descriptions for each motion part. The output requirements tell LLMs that we need structured output such as JSON format, body part naming, etc. Then, we employ a few-shot approach to guide LLMs in generating the desired output.
\revision{More details of our prompts can be found in the supplementary materials.}
A decomposed description example is shown in Table~\ref{tab:text_decomposition}.

\begin{table}
    \caption{An example of decomposing full-body motion description: \textit{``a person waves the right hand and then slightly bends down to the right and takes a few steps forward.''}}
    \label{tab:text_decomposition}
    \begin{tabular}{c|c}
        \toprule
        Part name & Part description          \\ \midrule
        head      & dose nothing              \\
        left arm  & dose nothing              \\
        right arm & waves hand                \\
        torso     & slightly bends down       \\
        left leg  & takes a few steps forward \\
        right leg & takes a few steps forward \\ \bottomrule
    \end{tabular}
\end{table}

\subsection{Part Motion Encoders}
The part motion encoders, $\{E_\mathrm{head}, \dots, E_\mathrm{right\_leg}\}$, aim to learn local semantic mapping from part-level input pairs $\left(\mathbf{M}_\mathrm{part}^n, T_\mathrm{part}\right)$ independently.
\revision{Since each encoder obtains information only from its corresponding part-level input pair and cannot access information from other body parts, the issue of semantic leakage is effectively alleviated.
}
We denote the part-level encoding process as follows:
\begin{equation}
    \mathbf{z}^n_\mathrm{part} = E_\mathrm{part}\left(\mathbf{M}^n_\mathrm{part}, T_\mathrm{part}, n\right),
\end{equation}
where each part motion encoder, $E_\mathrm{part}$, consists of three components: a linear layer, a text encoder, and a Conformer~\cite{gulati2020ConformerConvolutionaugmentedTransformer}. The linear layer aims to align the size of the latent dimension with that of the text encoder. We use six different frozen part-level TMR text encoders \cite{petrovich2023TMRTexttoMotionRetrieval}, each corresponding to one of the six body parts, which are pretrained on part-level motion-text pairs $\left(\mathbf{M}_\mathrm{part}, T_\mathrm{part}\right)$ respectively. Since the TMR model is trained only on motion description and motion data, and not on large visual datasets, the motion-related text embedding encoded by TMR is easier for the network to distinguish than that by CLIP. The projected motion and text embedding are then fused and processed by a Conformer\cite{gulati2020ConformerConvolutionaugmentedTransformer}. The Conformer incorporates convolution blocks into the Transformer \cite{vaswani2017AttentionAllYou} architecture to better capture temporal local features.
Moreover, previous work~\cite{alexanderson2023ListenDenoiseAction} shows the success of Conformer on music-to-dance task.

\begin{figure}
    \centering
    \includegraphics[width=\linewidth]{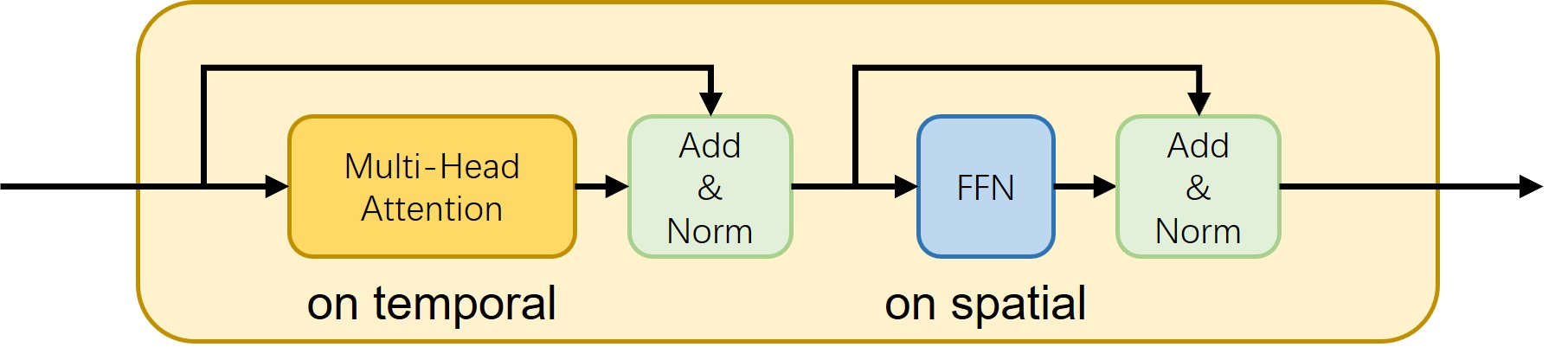}
    \caption{\revision{The structure of an attention encoder block.}}
    \label{fig:attention-encoder}
\end{figure}

\subsection{Full-Body Motion Optimizer}
Since each part's motion and text are independently encoded to $\left\{\mathbf{z}_\mathrm{head}^n, \cdots, \mathbf{z}_\mathrm{left\_leg}^n\right\}$ independently, the network will ignore the correlations between the different body parts, therefore, we propose that the full-body motion optimizer $G$ establishes correlations by adjusting the movements of each body part based on full-body text information.

Specifically, we first concatenate all body part latent codes into a full-body latent code $\mathbf{z}^n$ whose shape is $\left(F, S\right) = \left(F, 6 \times 128\right)$, and then fuse it with the global text embedding encoded by freezing the full-body level TMR text encoder. Next, we use an attention encoder~\cite{vaswani2017AttentionAllYou} to compute a delta that adjusts each part in the latent code $\mathbf{z}^n$.
\revision{The attention encoder is where the exchange of spatio-temporal information actually occurs. It consists of several attention encoder blocks, each containing a multi-head attention block and a feed-forward layer, as shown in Figure~\ref{fig:attention-encoder}. Since the latent code $\mathbf{z}^n$ is processed by a multi-head attention block on the temporal dimension $F$, and feed-forward layers (FFN) operate on the spatial dimension $S$, the latent code for each body part can continuously exchange temporal and spatial information.
    Next, we use a SmoothNet~\cite{zeng2022smoothnet} to reduce jitter, which contains a stacked MLP with residual connections and operates on the temporal dimension, acting as a low-pass filter in the latent space.
}

Finally, we project the latent code to origin feature dimension, and get a clean motion $\hat{\mathbf{M}}^0$. The full-body motion optimizer can be formulated as
\begin{equation}
    \begin{aligned}
        \hat{\mathbf{M}}^0 & = G\left(\mathbf{z}_\mathrm{head}^n, \cdots, \mathbf{z}_\mathrm{left\_leg}^n, T\right)                                   \\
                           & =  \mathrm{Linear}(\mathrm{SoothNet}(\mathbf{z}^n + \mathrm{AttentionEncoder}(\mathbf{z}_\mathrm{text} + \mathbf{z}^n)))
    \end{aligned}
\end{equation}
\section{Results}
In this section, we present the motions generated by our method and conduct a comparative analysis with other text-driven motion generation methods.
Additionally, we perform several ablation studies to highlight the contributions of individual components within our framework.

\begin{figure*}
    \centering
    \includegraphics[width=\linewidth]{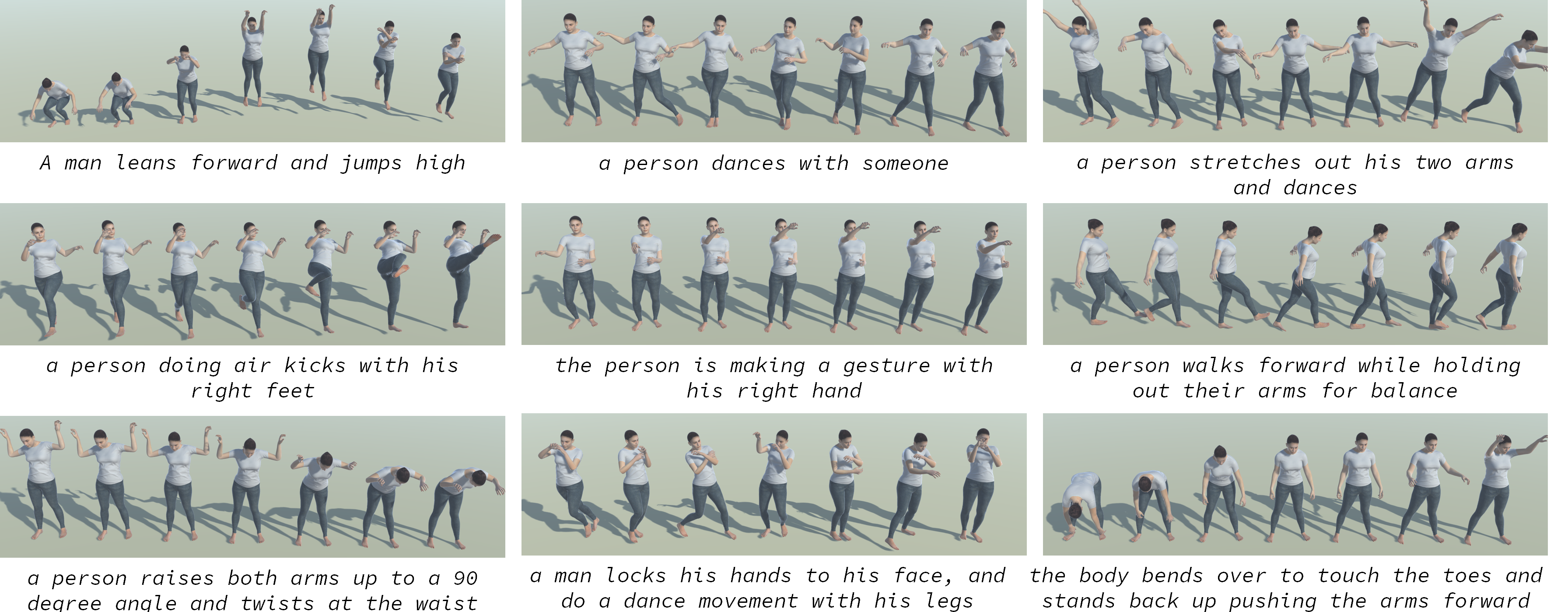}
    \caption[short]{Example results generated by our method.}
    \label{fig:gallery}
\end{figure*}

\subsection{Implementation Details}
The part-level motion description is generated by ChatGPT.  (\textit{gpt-3.5-turbo-1106}) model.
Our model is trained with AdamW optimizer with learning rate decaying strategy of fast warm cosine decay. The initial learning rate is $10^{-4}$ and the batch size is 64. The number of diffusion steps is 1K.
The training time of our model on the HumanML3D dataset is about 8 hours on 3 NVIDIA RTX 4090 GPUs.

\subsection{Qualitative Results}

Figure~\ref{fig:gallery} shows several example results generated by our method.
We can see that our method can generate motion with precise local semantics, such as body part semantic correspondence and action timing order, as our method injects local semantic information into corresponding parts independently, and the whole-body optimizer builds correct relationships between body parts in both spatial and temporal domains.
For example, the result of ``\textit{a man leans forward and jumps high}'' shows that the character does \textit{lean} and \textit{jump} in the correct order. The result of``\textit{a man lock his hands to his face, and do a dance move net with his legs}'' shows that the character keeps correct spatial relationship between hand and face while dancing. The result of ``\textit{a person doing air kicks with his right feet}'' shows that the character do \textit{kick} with correct body part.

\begin{figure*}
    \centering
    \includegraphics[width=\linewidth]{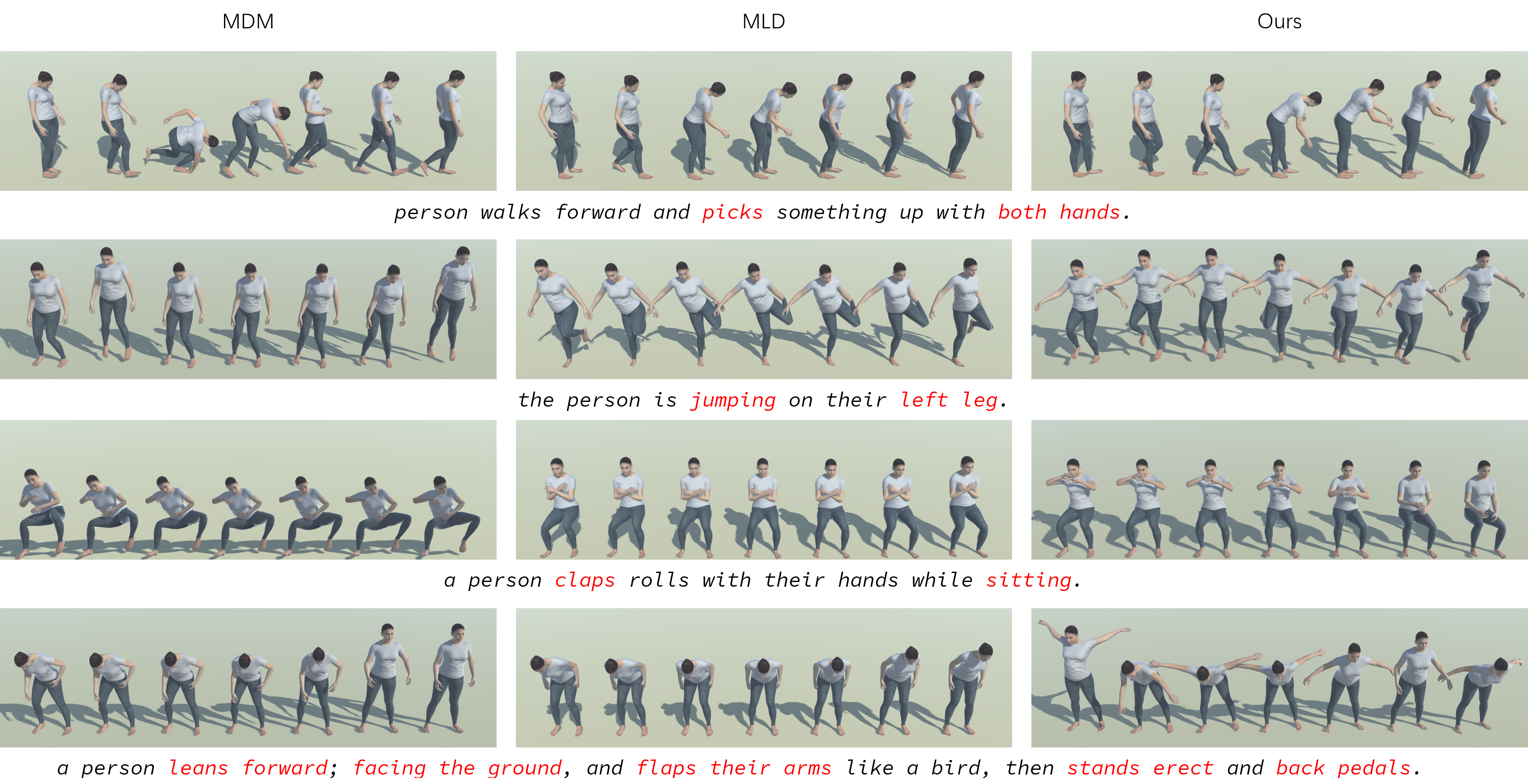}
    \caption{Qualitative comparison of results generated by our method with those from MDM~\cite{tevet2022HumanMotionDiffusion} and MLD~\cite{chen2023ExecutingYourCommandsa}.}
    \label{fig:comparison}
\end{figure*}

We also provide some visual comparisons to two baselines, including MDM~\cite{tevet2022HumanMotionDiffusion} and MLD~\cite{chen2023ExecutingYourCommandsa}.
Figure~\ref{fig:comparison} shows that our method can generate more semantic well-matched motion. In the first row, the character can pick something with \textit{both hands} in our result, but with just left hand in MDM. In the second row, the character only jumps on the left foot correctly in our result, but jumps on both feet in MDM and dose not jump in MLD. In the third row, the result of MDM contains weird pose and the MLD dose not contain ``claps'', but our result is more correct. The last row shows that, for more complex text inputs, our method is able to generate more semantic accurate results than those two baselines.

\subsection{Quantitative Evaluation}

\paragraph{Evaluation metrics.}
To quantitatively evaluate our method, we use the metrics suggested by \cite{guo2022GeneratingDiverseNatural} which includes
\begin{enumerate*}
    \item \textbf{Fréchet Inception Distance (FID)}  that evaluates the generated motion quality against real motion distribution;
    \item \textbf{Diversity (DIV)} that calculates the variance of generated motion;
    \item \textbf{R Precision} that calculates the top-n matching accuracy between generated motion and the corresponding text description;
    \item \textbf{Multi-Modal Distance (MM Dist)} that calculates the distance between paired motion and text;
    \item \textbf{Part-level Multi-Modal Similarity (PMM Sim)} that calculates the normalized cosine similarity between part-level paired motion and text.
\end{enumerate*}
These metrics are calculated in the latent space using the text encoder and motion encoder from T2M ~\cite{guo2022GeneratingDiverseNatural} as in previous works.
As our method provides detailed control of generated motions, we also compare our method to baselines in terms of part-level motion quality \revision{using Part-level Multi-Modal Similarity (PMM Sim)}, by training both part-level text encoder and motion encoder with contrastive learning as in TMR~\cite{petrovich2023TMRTexttoMotionRetrieval}, which we believe makes motion samples in the latent space more dispersed allowing dissimilar motions can be distinguished more easily.
\revision{Specifically, we calculate the PMM Sim in the TMR latent space as follows:}
\begin{equation}
    s_\mathrm{part} = \frac{1}{2}\left(\frac{z^{\mathrm{M}}_\mathrm{part}\cdot z^{T}_\mathrm{part}}{\|z^{\mathrm{M}}_\mathrm{part} \| \| z^{T}_\mathrm{part}\|}  + 1 \right)
\end{equation}
\revision{where both $z^{\mathrm{M}}_\mathrm{part}$ and $z^{T}_\mathrm{part}$ are obtained by encoding part-level motion and text through TMR encoders.
    Although we mainly focus on semantically controllable generation, we also evaluate common artifacts in text-to-motion synthesis. We assess the generated motions using three specific metrics: sliding, penetration, and floating, as introduced by \cite{yuan2022PhysDiffPhysicsGuidedHuman}.
}

\paragraph{Comparison results.}
The comparison results for full-body motion are presented in Tables~\ref{tab:full_body_metrics_t2m},
and the comparison results for part-level motion are presented in Table~\ref{tab:body_part_metrics}.
The FID and DIV in Tables~\ref{tab:full_body_metrics_t2m}
indicate that our method generates more realistic and diverse motion.
The R Precision and MM Dist indicate that our method can generate better globally semantically matching motion.
Table~\ref{tab:body_part_metrics} also shows that our method achieves the best local semantic matching, with performance very close to that of real data.
Our local-to-global design injects local semantic information independently into body parts and refines it with global semantics, which provides more accurate and structured semantic information to the network to help generation and thus achieve higher quality.
\revision{For artifact evaluation, as shown in Table~\ref{tab:artifacts_metrics}, we can see that each method exhibits performance very close to the ground truth (the Real row) at the millimeter scale. The artifacts can be attributed to the dataset's intrinsic quality variances. }

\begin{table}
    \centering
    \caption{Comparison of the visual quality and degree of semantic matching between input text and output full-body motion. These metrics are computed in the latent space of the T2M model~\cite{guo2022GeneratingDiverseNatural}.}
    \label{tab:full_body_metrics_t2m}
    \resizebox{\linewidth}{!}{
        \begin{tabular}{@{}l|cc|ccc|c@{}}
            \toprule
            \multirow{2}{*}{Method} & \multirow{2}{*}{FID $\downarrow$} & \multirow{2}{*}{DIV$\uparrow$} & \multicolumn{3}{c|}{R Precision$\uparrow$} & \multirow{2}{*}{MM Dist $\downarrow$} \\
            \cline{4-6}
                                                                             &                &                &         Top 1  &         Top 2  &         Top 3  &                \\\midrule
            Real                                                             &         0.000  &         9.831  &         0.513  &         0.708  &         0.804  &         2.939  \\ \midrule
            MotionDiffuse\citeyearpar{zhang2022MotionDiffuseTextDrivenHuman} &         0.687  &         8.894  &         0.318  &         0.531  &         0.677  &         3.118  \\
            MDM\citeyearpar{tevet2022HumanMotionDiffusion}                   &         0.747  &         9.462  &         0.390  &         0.581  &         0.695  &         3.635  \\
            MLD\citeyearpar{chen2023ExecutingYourCommandsa}                  &         1.753  &         8.970  &         0.383  &         0.573  &         0.687  &         3.682  \\ \midrule
            Ours (LGTM)                                                      & \textbf{0.218} & \textbf{9.638} & \textbf{0.490} & \textbf{0.689} & \textbf{0.788} & \textbf{3.013} \\ \bottomrule
        \end{tabular}
    }
\end{table}

\begin{table}
    \caption{Comparison of text-to-motion generation using \textbf{PMM Sim}. These metrics are calculated in the latent space of the part-level TRM encoder. Higher values indicate better performance.}
    \label{tab:body_part_metrics}
    \resizebox{\linewidth}{!}{
        \begin{tabular}{@{}l|cccccc@{}}
            \toprule
            Method           &          head  &       left arm  &     right arm  &         torso  &         left leg  &     right leg  \\ \midrule
            Real                                                             &         0.803  &          0.716  &         0.723  &         0.759  &            0.755  &         0.760  \\ \midrule
            MotionDiffuse\citeyearpar{zhang2022MotionDiffuseTextDrivenHuman} &         0.789  &          0.687  &         0.712  &         0.735  &            0.728  &         0.739  \\              
            MDM\citeyearpar{tevet2022HumanMotionDiffusion}                   &         0.783  &          0.699  &         0.691  &         0.740  &            0.717  &         0.723  \\
            MLD\citeyearpar{chen2023ExecutingYourCommandsa}                  &         0.771  &          0.675  &         0.702  &         0.717  &            0.723  &         0.726  \\ \midrule
            Ours (LGTM)                                                      & \textbf{0.799} &  \textbf{0.719} & \textbf{0.724} & \textbf{0.763} &    \textbf{0.755} & \textbf{0.763} \\ \bottomrule
        \end{tabular}
    }
\end{table}

\begin{table}
    \caption{\revision{Comparison of text to motion generation  using  metrics on artifact.}}
    \label{tab:artifacts_metrics}
    \resizebox{\linewidth}{!}{
        \begin{tabular}{@{}l|ccc@{}}
            \toprule
            Method & sliding (cm/s) $\downarrow$ & penetration (cm) $\downarrow$ & floating (cm) $\downarrow$ \\ \midrule
            Real                                                             &         0.743  &          1.442  &         0.079  \\ \midrule
            MotionDiffuse\citeyearpar{zhang2022MotionDiffuseTextDrivenHuman} &         1.359  &          1.783  &         0.051  \\              
            MDM\citeyearpar{tevet2022HumanMotionDiffusion}                   & \textbf{0.721} &          1.622  &         0.102  \\
            MLD\citeyearpar{chen2023ExecutingYourCommandsa}                  &         0.949  &          2.392  &         0.064  \\ \midrule
            Ours (LGTM)                                                      &         0.854  &  \textbf{1.247} & \textbf{0.046} \\ \bottomrule
        \end{tabular}
    }
\end{table}

\subsection{Ablation Studies}
We have designed two main experiments to assess the impact of different components of our approach. The first experiment investigates the influence of different text encoders on the motion quality. The second experiment evaluates the effect of the full-body motion optimizer on the the quality of motions generated by our method.

\paragraph{The importance of text encoder.}
We test our method by replacing our pre-trained text encoder with CLIP as an alternative, demonstrating that the TMR text encoder we use can capture more detailed semantics.
Furthermore, we also present the results obtained by MDM using either CLIP or the TMR text encoder for comparison.

Table~\ref{tab:text_encoder_ablation_global} and Table~\ref{tab:text_encoder_ablation_local} evaluate full-body and part-level motion quality, respectively.
In general, we observe that using the TMR text encoder consistently produces better results than using CLIP, for both our method and MDM as well as both local and global quality.
When comparing our method to MDM using the same text encoder, our method generally performs better, further demonstrating the superiority of our local-to-global design.

\begin{table}
    \caption{Comparison of the impact of different text encoders on full-body metrics computed in the latent space of T2M model~\cite{guo2022GeneratingDiverseNatural}.}
    \label{tab:text_encoder_ablation_global}
    \resizebox{\linewidth}{!}{
        \begin{tabular}{@{}l|cc|ccc|c@{}}
            \toprule
            \multirow{2}{*}{Method} & \multirow{2}{*}{FID $\downarrow$} & \multirow{2}{*}{DIV$\uparrow$} & \multicolumn{3}{c|}{R Precision$\uparrow$} & \multirow{2}{*}{MM Dist $\downarrow$} \\ 
            \cline{4-6}
                            &                 &                 &         Top 1  &         Top 2  &         Top 3  &                 \\ \midrule
            MDM + CLIP      &          0.747  &          9.462 &          0.390  &         0.581  &         0.695 &           3.635  \\
            MDM + TMR       &  \textbf{0.403} &  \textbf{9.687} & \textbf{0.455} & \textbf{0.653} & \textbf{0.759} &  \textbf{3.266} \\  \midrule
            Ours + CLIP     &          0.331 &           9.386  &         0.391  &         0.569  &         0.674  &          3.699  \\
            Ours + TMR      &  \textbf{0.218} &  \textbf{9.638} & \textbf{0.490} & \textbf{0.689} & \textbf{0.788} &  \textbf{3.013} \\ \bottomrule
        \end{tabular}
    }
\end{table}

\begin{table}
    \caption{Comparison of the impact of different text encoders on \textbf{PMM Sim} computed using the part-level TRM encoder. The greater the value, the better.}
    \label{tab:text_encoder_ablation_local}
    \resizebox{\linewidth}{!}{
        \begin{tabular}{@{}l|cccccc@{}}
            \toprule
            Method         &           head  &       left arm &     right arm &          torso  &      left leg  &     right leg  \\ \midrule
            MDM + CLIP     &          0.783  &         0.699  &         0.691  &         0.740  &         0.717  &         0.723  \\
            MDM + TMR      &  \textbf{0.803} & \textbf{0.704} & \textbf{0.707} & \textbf{0.756} & \textbf{0.734} & \textbf{0.743} \\ \midrule
            Ours + CLIP    &          0.795  &         0.693  &         0.694  &         0.752  &         0.725  &         0.732  \\ 
            Ours + TMR     &  \textbf{0.799} & \textbf{0.719} & \textbf{0.724} & \textbf{0.763} & \textbf{0.755} & \textbf{0.763} \\ \bottomrule
        \end{tabular}
    }
\end{table}

\revision{
    \paragraph{The impact of Conformer.}
    The goal of replacing Transformer with Conformer in Part Motion Encoders is to improve the motion quality. To validate the improvement, we compare both configurations on global quality metrics. From Table~\ref{tab:conformer_ablation_global} and Table~\ref{tab:conformer_ablation_local}, we observe that \sysname with Conformer can achieves better quality and semantic matching performance than with Transformer. This improvement can be attributed to the convolution blocks of Conformer, which capture local features better than self-attention.
}

\begin{table}
    \caption{\revision{Comparison of the impact of using Conformer versus Transformer in Part Motion Encoders on global quality.}}
    \label{tab:conformer_ablation_global}
    \resizebox{\linewidth}{!}{
        \begin{tabular}{@{}l|cc|ccc|c@{}}
            \toprule
            \multirow{2}{*}{Method} & \multirow{2}{*}{FID $\downarrow$} & \multirow{2}{*}{DIV$\uparrow$} & \multicolumn{3}{c|}{R Precision$\uparrow$} & \multirow{2}{*}{MM Dist $\downarrow$} \\ 
            \cline{4-6}
                           &                 &                 &         Top 1  &         Top 2  &         Top 3  &                 \\ \midrule
            Transformer    &          1.814  &          8.578 &          0.373  &         0.567  &         0.680  &          3.688 \\
            Conformer      &  \textbf{0.218} &  \textbf{9.638} & \textbf{0.490} & \textbf{0.689} & \textbf{0.788} &  \textbf{3.013} \\ \bottomrule
        \end{tabular}
    }
\end{table}

\begin{table}
    \caption{\revision{Comparison of the impact of using Conformer versus Transformer in Part Motion Encoders on {PMM Sim}. Higher values indicate better performance.}}
    \label{tab:conformer_ablation_local}
    \resizebox{\linewidth}{!}{
        \begin{tabular}{@{}l|cccccc@{}}
            \toprule
            Method      & head           & left arm       & right arm      & torso          & left leg       & right leg      \\ \midrule
            Transformer & 0.784          & 0.712          & 0.718          & 0.750          & 0.728          & 0.732          \\
            Conformer   & \textbf{0.799} & \textbf{0.719} & \textbf{0.724} & \textbf{0.763} & \textbf{0.755} & \textbf{0.763} \\ \bottomrule
        \end{tabular}
    }
\end{table}

\begin{figure}
    \centering
    \begin{subfigure}{0.48\linewidth}
        \includegraphics[width=\linewidth]{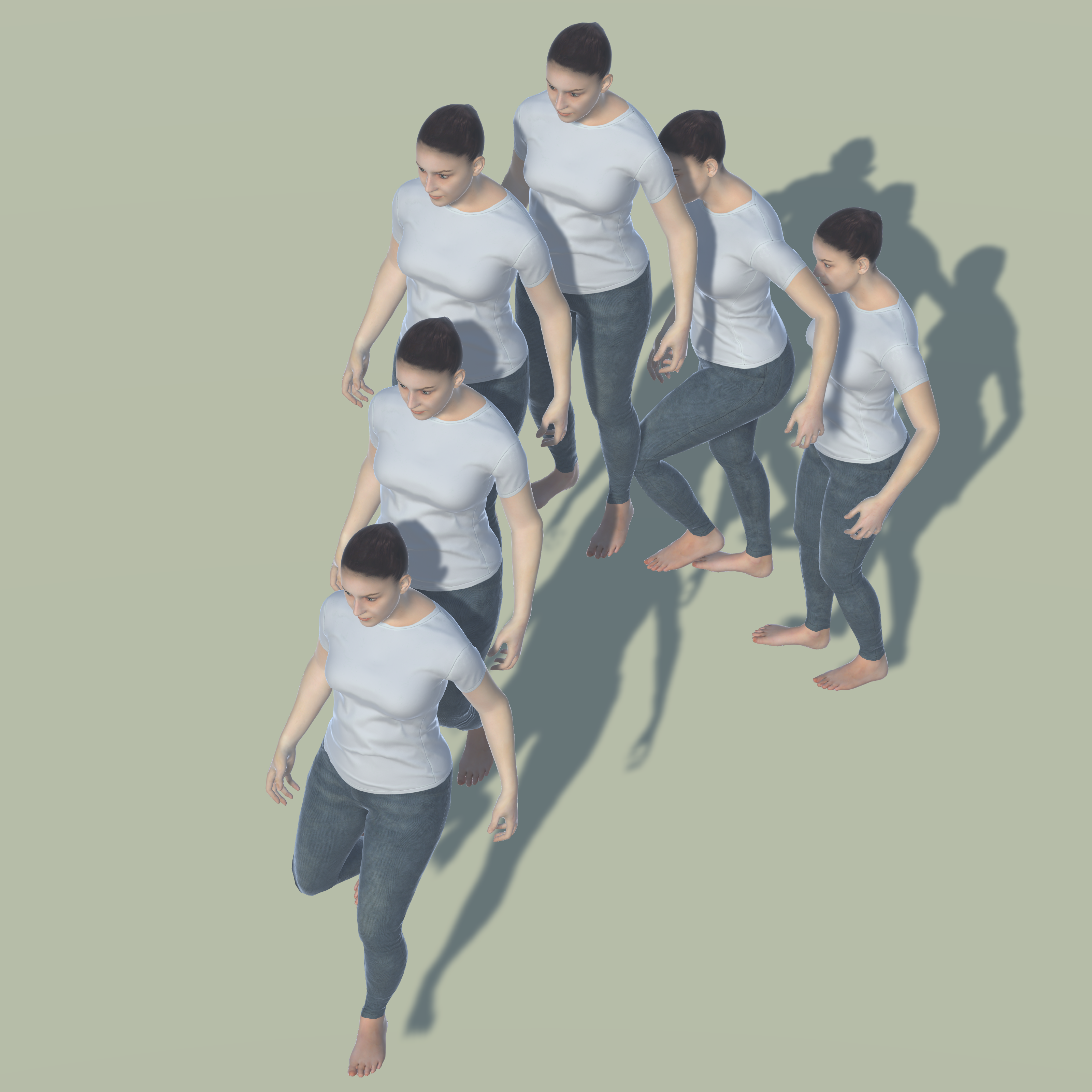}
        \caption{With full-body optimizer.}
    \end{subfigure}
    \begin{subfigure}{0.48\linewidth}
        \includegraphics[width=\linewidth]{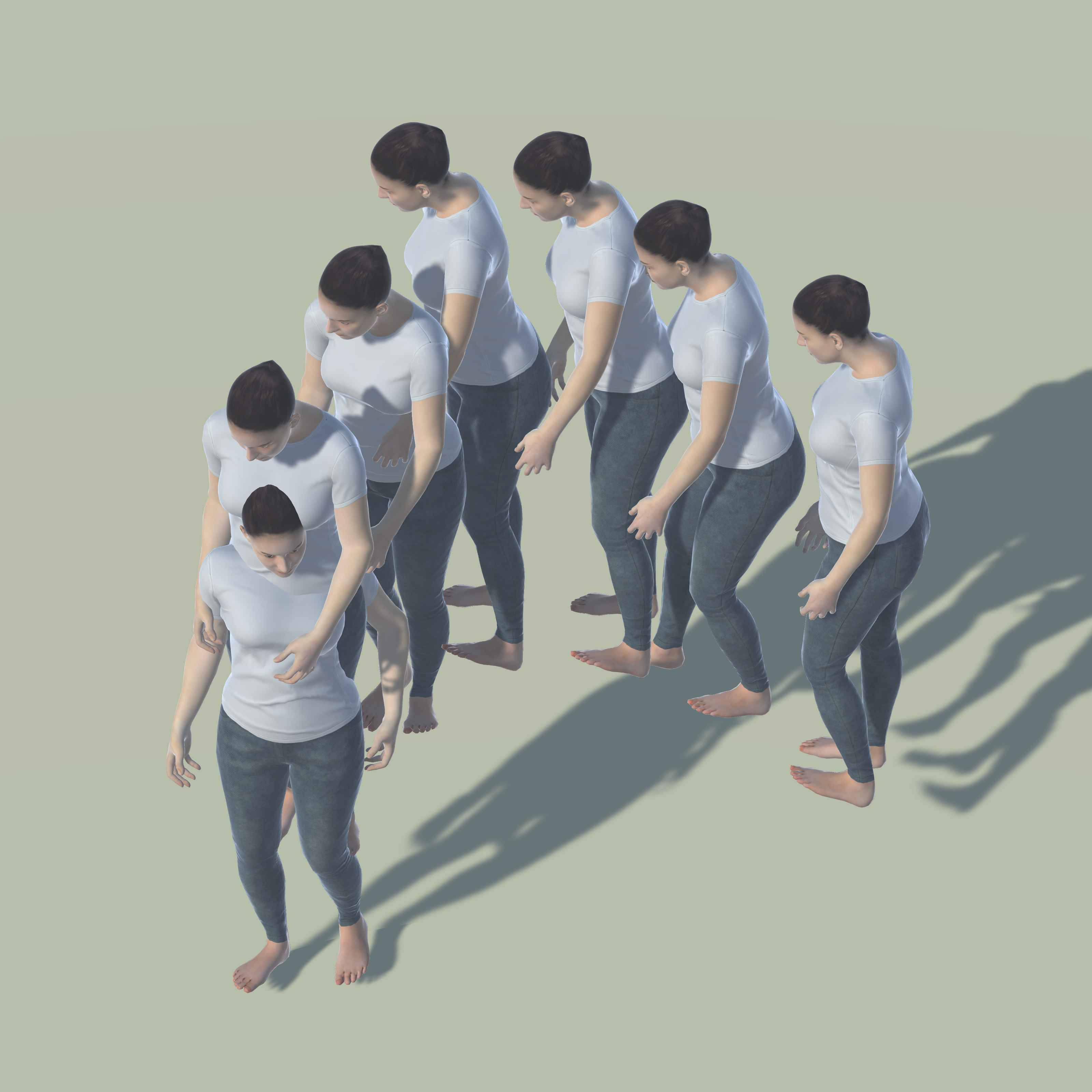}
        \caption{Without full-body optimizer.}
    \end{subfigure}
    \caption{Motions generation by our method with and without the full-body optimizer for \textit{``a person walks upstairs, turns left, and walks back downstairs.''}}
    \label{fig:ablation_optimizer}
\end{figure}

\paragraph{The importance of full-body motion optimizer.}
The goal of our full-body motion optimizer is to establish correlations among different body part movements and improve the coordination of full-body movements.
To validate the effect, we compare it to the setting ``w/o opt'', where we remove the key component of our full-body optimizer, namely, the attention encoder
From Table~\ref{tab:ablation_optimizer_global} and Table~\ref{tab:ablation_optimizer_local},
we can see that the local motion quality drops, and the full-body motion quality is also much worse without the optimizer; see Figure~\ref{fig:ablation_optimizer} for one example result.
Without the full-body optimizer, the character's two feet cannot coordinate well to step alternately during movement due to the lack of information exchange.

\begin{table}
    \caption{Comparison of the impact of attention encoder on global quality. These metric are calculated using the T2M model~\cite{guo2022GeneratingDiverseNatural}.}
    \label{tab:ablation_optimizer_global}
    \resizebox{0.45\textwidth}{!}{
        \begin{tabular}{@{}l|cc|ccc|c@{}}
            \toprule
            \multirow{2}{*}{Method} & \multirow{2}{*}{FID $\downarrow$} & \multirow{2}{*}{DIV$\uparrow$} & \multicolumn{3}{c|}{R Precision$\uparrow$} & \multirow{2}{*}{MM Dist $\downarrow$} \\ 
            \cline{4-6}
                     &                &                &         Top 1  &         Top 2  &         Top 3  &                 \\ \midrule
            w/o opt  &         7.384  &        11.552  &        0.219  &          0.360  &         0.454  &         5.227  \\ 
            w/ opt   & \textbf{0.218} & \textbf{9.638} & \textbf{0.490} & \textbf{0.689} & \textbf{0.788} & \textbf{3.013} \\ \bottomrule
        \end{tabular}
    }
\end{table}

\begin{table}
    \caption{Comparison of the impact of attention encoder on \textbf{PMM Sim} using the part-level TMR encoder. The greater the value, the better.}
    \label{tab:ablation_optimizer_local}
    \resizebox{0.48\textwidth}{!}{
        \begin{tabular}{@{}l|cccccc@{}}
            \toprule
            Method  &          head  &       left arm  &     right arm  &         torso  &         left leg  &     right leg  \\ \midrule
            w/o opt &         0.783  &          0.715  &         0.700  &         0.735  &            0.699  &         0.709  \\                                                           
            w/ opt  & \textbf{0.799} &  \textbf{0.719} & \textbf{0.724} & \textbf{0.763} &    \textbf{0.755} & \textbf{0.763} \\ \bottomrule
        \end{tabular}
    }
\end{table}

\section{Conclusion}

\begin{figure}
    \centering
    \includegraphics[width=\linewidth]{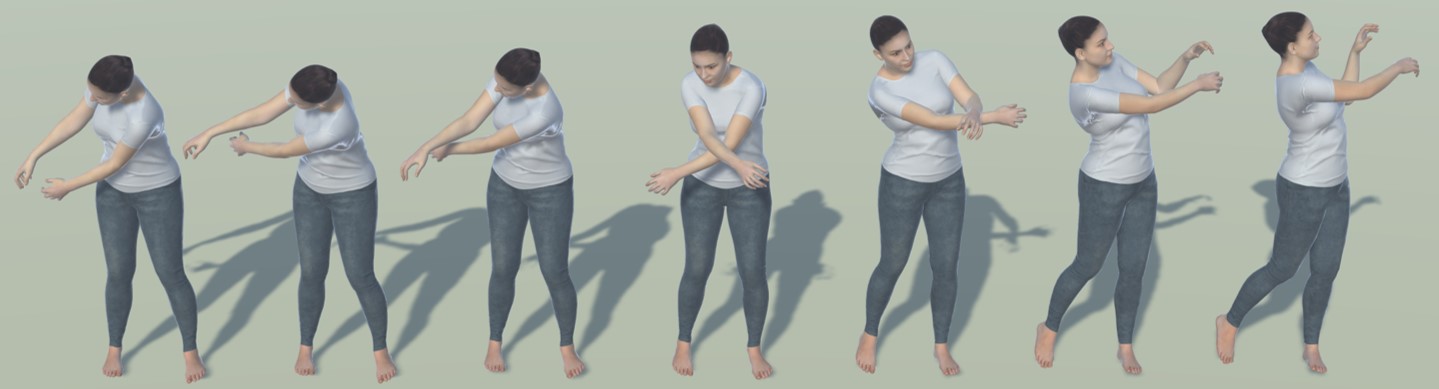}
    \caption{A failure case. The corresponding input prompt is \textit{someone imitating a golf swing.}}
    \label{fig:failure_case}
\end{figure}

In this study, we propose \sysname for text-to-motion generation, which significantly improves the accuracy and coherence of 3D human motion derived from textual descriptions. By integrating large language models with a local-to-global generation framework, our method effectively addresses key challenges in semantic mapping and motion coherence.

\paragraph{Limitation and future work.}

\revision{As we use ChatGPT for motion description decomposition, the local semantic mapping depends on the reasoning ability of ChatGPT. Incorrect decomposition or mapping may lead to unsatisfactory motion generation results. For example, when generating the \textit{``golf swing''} motion, which requires high-level and full-body coordination, \sysname struggles because ChatGPT identifies that the right hand swings the golf club but fails to decompose this reasoning into a series of low-level actions for each body part. The result is that the network generates an implausible motion, as shown in Figure~\ref{fig:failure_case}.
    Also, ambiguous texts in the dataset can confuse the network during training. For example, the phrase \textit{``a person performs action A and action B''} could imply that these actions occur simultaneously or sequentially, leading to output that may not align with user expectations.
    This issue could be mitigated by providing more detailed temporal descriptions.
    Furthermore, due to the limited length of samples in the dataset, our current framework cannot consistently generate long-term motions with high quality.
}
For future work, one promising direction is to incorporate our local-to-global idea with those VQ-VAE based approaches such as  TM2T~\cite{guo2022TM2TStochasticTokenized} and MotionGPT~\cite{jiang2023MotionGPTHumanMotion} by onstructing part-level motion clips as motion tokens for more detailed motion generation with different part-level motion combinations.

\begin{acks}
    We thank the anonymous reviewers for their valuable comments. This work was supported in parts by NSFC (62322207, 62161146005, U2001206), Guangdong Natural Science Foundation (2021B1515020085), Shenzhen Science and Technology Program (RCYX20210609103121030), DEGP Innovation Team (2022KCXTD025),  Guangdong Laboratory of Artificial Intelligence and Digital Economy (SZ) and Scientific Development Funds of Shenzhen University.
\end{acks}

\bibliographystyle{ACM-Reference-Format}
\bibliography{references}


\end{document}